\documentclass[nohyperref]{article}

\usepackage{microtype}
\usepackage{graphicx}
\usepackage{subfigure}
\usepackage{booktabs} 

\usepackage{hyperref}

\usepackage[accepted]{icml2023}

\usepackage{amsmath}
\usepackage{amssymb}
\usepackage{mathtools}
\usepackage{amsthm}

\usepackage[capitalize,noabbrev]{cleveref}

\theoremstyle{plain}

\theoremstyle{definition}

\theoremstyle{remark}

\usepackage[textsize=tiny]{todonotes}

\def\eqref#1{equation~\ref{#1}}

\def\1{\bm{1}}

\def\rvh{{\mathbf{h}}}

\def\rvv{{\mathbf{v}}}

\def\rvx{{\mathbf{x}}}
\def\rvy{{\mathbf{y}}}
\def\rvz{{\mathbf{z}}}

\def\ervh{{\textnormal{h}}}

\def\ervv{{\textnormal{v}}}

\def\vone{{\bm{1}}}

\def\vtheta{{\bm{\theta}}}

\def\vg{{\bm{g}}}
\def\vh{{\bm{h}}}

\def\vv{{\bm{v}}}

\def\vx{{\bm{x}}}
\def\vy{{\bm{y}}}
\def\vz{{\bm{z}}}

\def\evh{{h}}

\def\evv{{v}}

\def\mI{{\bm{I}}}

\def\mW{{\bm{W}}}

\DeclareMathAlphabet{\mathsfit}{\encodingdefault}{\sfdefault}{m}{sl}
\SetMathAlphabet{\mathsfit}{bold}{\encodingdefault}{\sfdefault}{bx}{n}

\def\emW{{W}}

\DeclareMathOperator*{\argmin}{arg\,min}

\usepackage{bm}

\icmltitlerunning{End-to-end Training of Deep Boltzmann Machines by Unbiased Contrastive Divergence with Local Mode Initialization}

\begin{document}

\twocolumn[
\icmltitle{End-to-end Training of Deep Boltzmann Machines \\by Unbiased Contrastive Divergence with Local Mode Initialization}



\icmlsetsymbol{equal}{*}

\begin{icmlauthorlist}
\icmlauthor{Shohei Taniguchi}{utokyo}
\icmlauthor{Masahiro Suzuki}{utokyo}
\icmlauthor{Yusuke Iwasawa}{utokyo}
\icmlauthor{Yutaka Matsuo}{utokyo}
\end{icmlauthorlist}

\icmlaffiliation{utokyo}{The University of Tokyo, 7-chōme-3-1 Hongō, Bunkyo City, Tokyo 113-8654, Japan}

\icmlcorrespondingauthor{Shohei Taniguchi}{taniguchi@weblab.t.u-tokyo.ac.jp}

\icmlkeywords{Machine Learning, ICML}

\vskip 0.3in
]



\printAffiliationsAndNotice{\icmlEqualContribution} 

\begin{abstract}
We address the problem of biased gradient estimation in deep Boltzmann machines (DBMs).
The existing method to obtain an unbiased estimator uses a maximal coupling based on a Gibbs sampler, but when the state is high-dimensional, it takes a long time to converge.
In this study, we propose to use a coupling based on the Metropolis-Hastings (MH) and to initialize the state around a local mode of the target distribution.
Because of the propensity of MH to reject proposals, the coupling tends to converge in only one step with a high probability, leading to high efficiency.
We find that our method allows DBMs to be trained in an end-to-end fashion without greedy pretraining. We also propose some practical techniques to further improve the performance of DBMs. 
We empirically demonstrate that our training algorithm enables DBMs to show comparable generative performance to other deep generative models, achieving the FID score of $10.33$ for MNIST.

\end{abstract}

\section{Introduction}
\label{sec:intro}
Boltzmann machines \citep{fahlman1983massively,ackley1985learning,hinton1984boltzmann,hinton1986learning}, which are energy-based models defined over binary vectors, have played an important role in the history of deep learning. For example, restricted Boltzmann machines (RBMs), Boltzmann machines with a single hidden layer, have been used for the pretraining of feedforward neural networks (FNNs)~\citep{hinton2006fast,hinton2006reducing,hinton2007recognize,bengio2006greedy,ranzato2006efficient}.
Although Boltzmann machines are rarely used for pretraining today, they have many applications as undirected generative models, e.g.,  missing value imputation of incomplete data~\citep{wang2017shape,ma2020restricted}, multimodal learning~\citep{srivastava2012multimodal}, and collaborative filtering~\citep{salakhutdinov2007restricted}.
Boltzmann machines also have the potential as powerful generative models because it is known as a universal approximator of the probability mass function on discrete variables~\citep{le2008representational}.
Among them, deep Boltzmann machines (DBMs)~\citep{salakhutdinov2010efficient}, which are multi-layered undirected models, can capture complex structures by their deep structure while retaining the advantages of the Boltzmann machine.

However, the training of DBMs is known to be quite difficult.
The difficulty in training is largely related to the gradient estimation in DBMs.
Maximum likelihood learning of DBMs requires to estimate the gradient of their marginal log-likelihood.
However, this gradient estimator tends to be biased because it involves several approximations.
Specifically, this gradient estimator includes the expectation over the posterior and their joint distribution. 
Since these distributions are intractable in DBMs, some approximations are required, such as a variational approximation (for the posterior) and a persistent Gibbs sampler (for the joint distribution)~\citep{salakhutdinov2010efficient}.
Such approximations typically lead to biased gradient estimations.
Both theoretical and empirical studies have shown that biased gradient estimators harm the convergence property of stochastic approximation algorithms~\citep{carreira2005contrastive,schulz2010investigating,fischer2010empirical}.
Therefore, it is necessary to seek an unbiased gradient estimator to mitigate the learning difficulties of DBM.
Recently, \citet{jacob2020unbiased} have established a theory of an unbiased Markov chain Monte Carlo (MCMC) method that removes the bias of an MCMC-based estimator using a technique called {\it couplings}.
The unbiased MCMC method enables us to construct an unbiased estimator of some expectation over an intractable random variable in a finite time.
Based on this theory, \citet{Qiu2020Unbiased} have proposed the unbiased contrastive divergence (UCD) algorithm, in which an unbiased gradient estimator of the marginal log-likelihood for energy-based latent variable models like RBMs and DBMs is constructed using a Gibbs-sampler-based coupling.
They have shown that the UCD algorithm is empirically effective in avoiding non-convergent behavior in the training of RBMs. 
However, we empirically find that their Gibbs-sampler-based algorithm tends to take a long time to converge, especially in high-dimensional cases, due to the curse of dimensionality.
In fact, their experiments are conducted only with relatively low-dimensional RBMs; therefore, the difficulty of applying UCD to DBMs, which tend to be more high-dimensional than RBMs, still remains.
To use the UCD algorithm for unbiased estimation in DBMs, we need to establish a method where couplings converge in less time, even in high dimensions.

In this paper, we propose an improved UCD algorithm to efficiently perform unbiased gradient estimation in DBMs.
The main insight of the proposed method is to search for an initial state of the Markov chain that reduces the number of coupling steps in UCD.
To achieve this, we first change the Markov chain of couplings in UCD from a Gibbs sampler to the Metropolis-Hastings (MH) algorithm.
The MH algorithm has a property that it rejects a proposal for the next state with a high probability if the energy of the current state is low.
Therefore, we propose to set the initial value of the Markov chain to a state with low energy, i.e., near a \emph{local mode}.
This allows us to terminate the Markov chain to obtain an unbiased gradient estimator in just a few steps.
More importantly, if the state is high-dimensional, the acceptance probability in the MH is even lower, resulting in termination with fewer coupling steps, such as $1$.
This means that the proposed method greatly alleviates the problem of the curse of dimensionality in the UCD algorithm.

To explore a local mode, we use the local search algorithm~\citep{hromkovivc2013algorithmics}, in which the state is iteratively updated from a random initialization by applying local changes to the state until convergence.
Although a certain number of steps is additionally required to perform the local search, the proposed method is much more efficient in practice than the original UCD.
We name this proposed learning algorithm of DBMs the {\it unbiased contrastive divergence with local mode initialization} (UCD-LMI).

Furthermore, we find that UCD-LMI allows us to learn DBMs without pretraining.
DBMs are known to fail to train from random initialization, which is called the joint training problem~\citep{Goodfellow-et-al-2016}, and a widely known solution to this is greedy layer-wise pretraining.
Although there have been attempts to train without pertaining~\citep{montavon2012deep,goodfellow2013multi}, it is still difficult to show good generative performance. Our method shows high-generation performance by end-to-end training without pretraining and thus can be considered a potential solution to the joint training problem. 


We also propose several practical techniques, including a simplified centering trick, a marginalization trick to reduce the variance, and orthogonal initialization,  needed to successfully train a DBM.
We empirically find that these additional techniques are also essential to achieve performance comparable to other DGMs.

In the experiment, we compare our method to existing learning methods of DBMs and RBMs on the image generation benchmarks using the MNIST, Fashion-MNIST, and CIFAR-10 datasets.
Our method achieves the FID score of $10.33$ for MNIST, which is comparable to the Wasserstein generative adversarial network (WGAN). 


\section{Preliminaries}
\label{sec:review}

\subsection{Deep Boltzmann Machine}

The deep Boltzmann Machine (DBM)~\citep{salakhutdinov2010efficient} is a kind of Boltzmann machine, which has several layers of latent variables.
In the case of a deep Boltzmann machine with a visible layer, $\rvv$, and two hidden layers, $\rvh^{(1)}$ and $\rvh^{(2)}$, the joint probability is given by
\begin{align}
    P\left(\rvv, \rvh^{(1)}, \rvh^{(2)} \right) \propto \exp \left(-E (\rvv, \rvh^{(1)}, \rvh^{(2)} ; \vtheta )\right).
\end{align}
Each unit of $\rvv$, $\rvh^{(1)}$, and $\rvh^{(2)}$ takes a binary value of $0$ or $1$.
The energy function $E$ is defined as follows:
\begin{align}
    &E (\rvv, \rvh^{(1)}, \rvh^{(2)} ; \vtheta) \nonumber \\
    &= - \rvv^\top \mW^{(1)} \rvh^{(1)} - {\rvh^{(1)}}^\top \mW^{(2)} \rvh^{(2)}, \label{eq:energy}
\end{align}
where $\vtheta$ denotes all parameters, i.e., $\vtheta = \left\{ \mW^{(1)}, \mW^{(2)} \right\}$, and the bias parameters are omitted in this definition for clarity.
Because the DBM has a bipartite structure, we can easily calculate the conditional distribution over the odd layers and the even layers as follows:
\begin{gather}
    P ( \ervv_i = 1 \mid \vh^{(1)} ) = \sigma \left( \mW_{i,:}^{(1)} \vh^{(1)} \right), \\
    P ( \ervh_j^{(1)} = 1 \mid \vv, \vh^{(2)} ) = \sigma \left( \vv^\top \mW_{:,j}^{(1)} + \mW_{j,:}^{(2)} \vh^{(2)} \right), \\
    P ( \ervh_k^{(2)} = 1 \mid \vh^{(1)} ) = \sigma \left( {\vh^{(1)}}^\top \mW_{k,:}^{(2)} \right).
\end{gather}
We here consider a DBM with two hidden layers for simplicity, but it can be easily extended to deeper models.

Although the Boltzmann machine (including the DBM) is originally developed for binary variables, it can be extended to the case where the visible units are real-valued by slightly modifying the energy function~\citep{welling2004exponential,hinton2006reducing}.
This extension is called the Gaussian-Bernoulli Boltzmann machine because its conditional distribution over the visible layer is a Gaussian.


\subsection{Training DBM}
To learn a DBM by maximum likelihood, we need to calculate the gradient of the log-likelihood with respect to the parameters $\vtheta$ as follows.
\begin{gather}
\begin{aligned}
    \nabla_\vtheta \log p ( \vv ; \vtheta ) = &- \mathbb{E} \left[ \nabla_\vtheta E (\vv, \rvh^{(1)}, \rvh^{(2)} ; \vtheta ) \right] \\
    &+ \mathbb{E} \left[ \nabla_\vtheta E (\tilde{\rvv}, \tilde{\rvh}^{(1)}, \tilde{\rvh}^{(2)} ; \vtheta )\right],
\end{aligned} \label{eq:gradient}\\
    \rvh^{(1)}, \rvh^{(2)} \sim P \left( \rvh^{(1)}, \rvh^{(2)} \mid \vv \right), \\
    \tilde{\rvv}, \tilde{\rvh}^{(1)}, \tilde{\rvh}^{(2)} \sim P \left( \rvv, \rvh^{(1)}, \rvh^{(2)} \right).
\end{gather}
However, the expectations with respect to the posterior distribution over the hidden layers $P \left( \rvh^{(1)}, \rvh^{(2)} \mid \vv \right)$ and the joint distribution $P \left( \rvv, \rvh^{(1)}, \rvh^{(2)} \right)$ are both intractable, so we need some approximation.
A common way is to use a variational approximation for the posterior and a persistent Gibbs sampler for the joint distribution respectively.
In the variational approximation, the posterior $P \left( \rvh^{(1)}, \rvh^{(2)} \mid \vv \right)$ is approximated by a variational approximate distribution $Q \left( \vh^{(1)}, \vh^{(2)} \mid \vv \right)$, and $Q$ is also optimized by minimizing the Kullback–Leibler divergence to the true posterior.

The expectation over the joint distribution is approximated using samples obtained by running some steps of a Gibbs sampling.
The Gibbs chain at each gradient step is initialized with their states from the previous gradient step.
This approach is called {\it stochastic maximum likelihood} (SML)~\citep{younes1998stochastic} or {\it persistent contrastive divergence} (PCD)~\citep{tieleman2008training}.

\subsection{Joint Training Problem}
Unfortunately, training a DBM using PCD from a random initialization usually results in failure.
This problem is often called the {\it joint training problem}~\citep{Goodfellow-et-al-2016}.
The most popular method for overcoming the joint training problem is greedy layer-wise pretraining, in which each layer of the DBM is trained in isolation as an RBM.
%
Although greedy pretraining is useful for training a DBM properly, it has some undesirable properties, e.g., the complexity of software implementations and the difficulty of tracking the performance during training.

To address them, there have been some attempts to train a DBM without pretraining.
For example, the centered DBM~\citep{montavon2012deep} uses a centering trick, in which the unit values are normalized using moving statistics, enabling accelerated training of DBMs.
The multi-prediction DBM~\citep{goodfellow2013multi} is another example of DBMs that does not require greedy pretraining, in which the model is trained using the back-propagation algorithm without relying on Markov chain Monte Carlo (MCMC) like Gibbs sampling.
Although these methods can alleviate the joint training problem, it is still difficult to achieve good generative performance with DBMs.



What makes the training of DBMs more difficult than RBMs is that the posterior distribution over all the hidden units $P (\rvh \mid \rvv )$ is intractable for DBMs, whereas it has an analytic form for RBMs.
Therefore, training an RBM is much easier than a DBM because the first term of Eq. (\ref{eq:gradient}) can be easily approximated using exact samples from the posterior.
In the case of a DBM, on the other hand, the posterior is no longer tractable; hence an approximation has to be performed.
However, since existing approximation methods (e.g., variational approximation) have bias in their estimation, the resulting algorithm tends not to converge even to local minima of the log-likelihood, leading to poor performance.
The problem of biased estimation applies not only to the posterior $P \left( \vh^{(1)}, \vh^{(2)} \mid \vv \right)$ but also to the joint distribution $P \left( \vv, \vh^{(1)}, \vh^{(2)} \right)$, because PCD also has bias as an estimator.
We hypothesize that the lack of unbiasedness in the gradient estimator is a critical issue that causes difficulty in training DBMs.

\section{Unbiased Gradient Estimation}
\label{sec:method}
In this section, we address the joint training problem of DBMs by proposing an algorithm that removes the bias of gradient estimation, enabling stable training of DBMs. 
First, we introduce a general concept of the unbiased MCMC method. 
Subsequently, we explain an existing UCD algorithm, which uses the unbiased MCMC for the training of DBMs, and point out its problem.
Finally, our UCD-LCI algorithm is derived by addressing the problem. 
\subsection{Unbiased MCMC with Couplings}
\citet{jacob2020unbiased} have proposed an unbiased MCMC method to remove the bias of MCMC-based estimators.
Suppose we want to estimate the expectation over some function $f$ in terms of a random variable $\rvx \sim P \left( \rvx \right)$, i.e.,  $\mathbb{E} \left[ f \left( \rvx \right) \right]$. If a Markov chain $\{\rvx_t \}$ satisfies $\mathbb{E} \left[ f \left( \rvx_t \right) \right] \rightarrow \mathbb{E} \left[ f \left( \rvx \right) \right]$ as $t \to \infty$, then under some regularity conditions, the expectation can be expressed as a telescoping sum as follows:
\begin{align}
    \mathbb{E} \left[ f \left( \rvx \right) \right] = \mathbb{E} \left[ f \left( \rvx_0 \right) + \sum_{t=1}^\infty f \left( \rvx_t \right) - f \left( \rvx_{t-1} \right) \right].
\end{align}
In addition, if we assume that there is another Markov chain $\{\rvy_t \}$ such that (1) $\rvx_t$ and $\rvy_t$ have the same marginal distributions for all $t \geq 0$, and (2) $\rvx_t = \rvy_{t-1}$ for all $t \geq \tau \geq 1$, where $\tau$ is some random time, the following equation holds:
\begin{align}
    \mathbb{E} \left[ f \left( \rvx \right) \right] = \mathbb{E} \left[ f \left( \rvx_0 \right) + \sum_{t=1}^{\tau-1} f \left( \rvx_t \right) - f \left( \rvy_{t-1} \right) \right].
\end{align}
Therefore, the quantity $f \left( \rvx_0 \right) + \sum_{t=1}^{\tau-1} f \left( \rvx_t \right) - f \left( \rvy_{t-1} \right)$ is an unbiased estimator for the expectation $\mathbb{E} \left[ f \left( \rvx \right) \right]$.
By carefully designing a coupling of the two Markov chains $\{ (\rvx_t, \rvy_t) \}$, we can obtain an unbiased estimator in a finite time $\tau$.
We refer to the random time $\tau$ as the {\it coupling time}.

A key ingredient of the unbiased MCMC algorithm is the construction of the coupling $\{ (\rvx_t, \rvy_t) \}$.
A common way to construct such Markov chains is to use a maximal coupling as provided by \citet{jacob2020unbiased}, where the probability of $P (\rvx_{t} = \rvy_{t-1} \mid \rvx_{t-1}, \rvy_{t-2})$ is maximized at each step $t \geq 2$.
By using the maximal coupling, the coupling time can be shortened; therefore, the unbiased estimator can be obtained efficiently.

\subsection{Unbiased Contrastive Divergence}
The unbiased MCMC method can be directly applied to the gradient estimation of a DBM in Eq. (\ref{eq:gradient}).
For example, an unbiased estimator of the second term of Eq. (\ref{eq:gradient}) can be obtained as follows:
\begin{align}
    &\mathbb{E} \left[ \nabla E ( \rvx ; \vtheta ) \right]  \label{eq:unbiased_energy_grad}\\
    &= \mathbb{E} \left[ \nabla E ( \rvx_0 ; \vtheta ) + \sum_{t=1}^{\tau-1} \nabla E ( \rvx_t ; \vtheta ) - \nabla E ( \rvy_{t-1} ; \vtheta ) \right], \nonumber
\end{align}
where $\rvx = (\tilde{\rvv}, \tilde{\rvh}^{(1)}, \tilde{\rvh}^{(2)})$, and $\{ (\rvx_t, \rvy_t) \}$ is a coupling of Markov chains that satisfies (1) $\mathbb{E} \left[ \nabla E \left( \rvx_t; \vtheta \right) \right] \to \mathbb{E} \left[ \nabla E \left( \rvx; \vtheta \right) \right]$, (2) $\rvx_t$ and $\rvy_t$ have the same marginal distributions for all $t \geq 0$, and (3) $\rvx_t = \rvy_{t-1}$ after some random time $\tau$.
A similar derivation can be applied to the first term of Eq. (\ref{eq:gradient}) (see Appendix \ref{sec:unbiased_first_term}).

\citet{Qiu2020Unbiased} have proposed a method to construct a maximal coupling based on a Gibbs sampler for energy-based latent variable models such as DBMs, and named the resulting learning algorithm the {\it unbiased contrastive divergence} (UCD).
However, we empirically observe that the coupling time of the Gibbs-based coupling tends to be very long, especially when $\rvx$ is high-dimensional due to the curse of dimensionality.
Therefore, it is difficult to apply this method to high-dimensional DBMs.


\subsection{UCD with Local Mode Initialization}
\label{subsec:ucd-lmi}
To apply the UCD algorithm to DBMs, we develop an efficient coupling algorithm to construct an unbiased gradient estimator of the log-likelihood in Eq. (\ref{eq:gradient}).
Here, we focus on the construction of an unbiased estimator for the second term of Eq. (\ref{eq:gradient}), but a similar derivation can be applied to the first term (see Appendix~\ref{sec:unbiased_first_term} for more details).


Our proposed coupling is based on the Metropolis–Hastings (MH) algorithm.
A Markov chain of MH is constructed using a proposal distribution $Q$ as follows:
\begin{gather}
    P \left( \rvx_t \mid \rvx_{t-1} \right) = Q \left( \rvx_t \mid \rvx_{t-1} \right) A \left( \rvx_t, \rvx_{t-1} \right), \\
    A \left( \rvx_t, \rvx_{t-1} \right) = \min \left( 1, \frac{P ( \rvx_t )}{P ( \rvx_{t-1} )} \frac{Q ( \rvx_{t-1} \mid \rvx_t )}{Q ( \rvx_{t} \mid \rvx_{t-1} )} \right).
\end{gather}
To obtain samples from this Markov chain, we first initialize the sample $\vx_0$, and then repeat iterations of generating a proposal $\vx^\prime$ according to $Q ( \rvx_t \mid \vx_{t-1} )$ and setting $\vx_t = \vx^\prime$ with probability $A \left( \vx^\prime, \vx_{t-1} \right)$, otherwise $\vx_t = \vx_{t-1}$.
In our method, we simply use a uniform distribution for the proposal $Q$.
In this case, the acceptance probability is simplified as follows:
\begin{align}
    A \left( \rvx_t, \rvx_{t-1} \right) 
    &= \min \left( 1, {P ( \rvx_t )} / {P ( \rvx_{t-1} )} \right) \\
    &= \min \left( 1, \exp ( E ( \rvx_{t-1} ; \vtheta ) - E ( \rvx_t ; \vtheta ) ) \right) \nonumber
\end{align}
In fact, a maximal coupling of the MH algorithm with the uniform proposal distribution can be easily constructed as shown in Algorithm \ref{alg:maximal_coupling} in the appendix~\citep{wang2021maximal}.
One might think that this coupling is less efficient than the Gibbs-based maximal coupling because the MH algorithm tends to mix slower than the Gibbs sampler, especially when the proposal is uniformly distributed.
However, if we set $\rvx_0 = \rvy_0$, and its energy $E ( \rvx_0 ; \vtheta )$ takes a very low value, the proposal for $\rvx_1$ is rejected with a very high probability, resulting in $\tau = 1$.
This means that by starting from a low-energy state, this MH-based coupling can be very efficient, because the coupling time would be almost always $1$.
Importantly, this tendency is more pronounced when $\vx$ is high-dimensional, because the acceptance probability of the MH step gets even lower in such cases.
In other words, this method turns the curse of dimensionality in the MH algorithm into an advantage to shorten the coupling time.

A challenge to efficiently perform MH-based maximal coupling is how to find low-energy states for its initialization.
To address it, we propose to use a local search algorithm~\citep{battiti2008reactive}.
Because the DBM has a bipartite structure as described in Section \ref{sec:review}, the conditional minimum of the energy function for the even layers given the odd layers can be easily obtained as follows:
\begin{align}
    &\argmin_{\rvv} \left\{ E \mid \rvh^{(1)} \right\} = \vone_{\mW^{(1)} \rvh^{(1)} \geq 0} \\
    &\argmin_{\rvh^{(2)}} \left\{ E \mid \rvh^{(1)} \right\} = \vone_{{\rvh^{(1)}}^\top \mW^{(2)} \geq 0}
\end{align}
Similarly, the conditional minimum for the odd layers given the even layers is as follows:
\begin{align}
    &\argmin_{\rvh^{(1)}} \left\{ E \mid \rvv, \rvh^{(1)} \right\} = \vone_{\rvv^\top \mW^{(1)} + \mW^{(2)} \rvh^{(2)} \geq 0}
\end{align}
As in Gibbs sampling, we iteratively update a sample of $\rvx = (\tilde{\rvv}, \tilde{\rvh}^{(1)}, \tilde{\rvh}^{(2)})$ by finding the conditional minimum of each block until the sample converges to a local minimum of the energy function, which is equivalent to a local mode of the target distribution $P ( \rvv, \rvh ; \vtheta )$.
A concrete algorithm of the local search for DBMs is provided in Algorithm \ref{alg:local_search}.
Note that the uniform noise $u$ is used to determine whether to update the even or odd layers first.
This local search algorithm converges in finite steps, because the number of possible states is countably finite in binary cases.
After finding a local mode, we run a single step of Gibbs sampling from it, and use the moved state as an initial state of Algorithm \ref{alg:maximal_coupling}.
The reason for running a single Gibbs step is to prevent the support of the initial state distribution of Algorithm \ref{alg:maximal_coupling} from being limited only to the local modes.
By this local mode initialization, it can be expected that the initial state of Algorithm \ref{alg:maximal_coupling} has low energy, making the coupling time $\tau$ almost always $1$, especially when $\vx$ is high-dimensional.

Using the local search to initialize the MH-based coupling yields an efficient learning algorithm of DBMs, which we name the {\it unbiased contrastive divergence with local mode initialization} (UCD-LMI).
The UCD-LMI is summarized in Algorithm \ref{alg:ucd-lmi}.
Although the gradient is estimated using a single training example for simplicity, it is possible to use a minibatch of some training examples, and average the estimates to reduce the variance as usual.
The parameter update rule of the simple stochastic gradient ascent can also be replaced by another fancy optimizer like Adam~\citep{kingma2014adam} and AMSGrad~\citep{reddi2019convergence}.

\begin{algorithm}[tb]
   \caption{Local Search Algorithm for DBM}
   \label{alg:local_search}
\begin{algorithmic}
   \STATE Draw an initial state $\left( \vv_0, \vh_0^{(1)}, \vh_0^{(2)} \right)$ uniformly.
   \STATE Draw $u$ uniformly from $[0, 1)$. 
   \STATE $t \leftarrow 0$
   \REPEAT
   \IF{$u < 0.5$}
   \STATE $\vv_{t+1} = \vone_{\mW^{(1)} \vh_t^{(1)} \geq 0}, \ \vh^{(2)}_{t+1} = \vone_{{\vh_t^{(1)}}^\top \mW^{(2)} \geq 0}$
   \STATE $\vh^{(1)}_{t+1} = \vone_{\vv_{t+1}^\top \mW^{(1)} + \mW^{(2)} \vh_{t+1}^{(2)} \geq 0}$
   \ELSE
   \STATE $\vh_{t+1}^{(1)} = \vone_{\vv_t^\top \mW^{(1)} + \mW^{(2)} \vh_t^{(2)} \geq 0}$
   \STATE $\vv_{t+1} \leftarrow \vone_{\mW^{(1)} \vh_{t+1}^{(1)} \geq 0}, \ \vh_{t+1}^{(2)} \leftarrow \vone_{{\vh_{t+1}^{(1)}}^\top \mW^{(2)} \geq 0}$
   \ENDIF
   \STATE $t \leftarrow t + 1$
   \UNTIL{$\left( \vv_t, \vh_t^{(1)}, \vh_t^{(2)} \right) = \left( \vv_{t-1}, \vh_{t-1}^{(1)}, \vh_{t-1}^{(2)} \right)$}
   \STATE $T \leftarrow t$
   \STATE {\bfseries return} $\left( \vv_T, \vh_T^{(1)}, \vh_T^{(2)} \right)$, $T$
\end{algorithmic}
\end{algorithm}

\begin{algorithm}[tb]
   \caption{UCD-LMI Algorithm}
   \label{alg:ucd-lmi}
\begin{algorithmic}
   \STATE Draw an initial parameter $\vtheta$.
   \REPEAT
   \STATE Sample an example $\vv$ from the training set.
   \STATE Run Algorithm \ref{alg:local_search_for_posterior}, and obtain the final value $\vh_T^{(1)}, \vh_T^{(2)}$.
   \STATE Run a single Gibbs step from $\vh_T^{(1)}, \vh_T^{(2)}$, and obtain $\vh_{T+1}^{(1)}, \vh_{T+1}^{(2)}$.
   \STATE Set $\vh_0 = \vz_0 = ( \vh_{T+1}^{(1)}, \vh_{T+1}^{(2)} )$.
   \STATE Run Algorithm \ref{alg:maximal_coupling_for_posterior} , and obtain $\{ (\vh_{t}, \vz_{t-1}) \}, \ \tau$.
   \STATE Set $\{ \vx_t \} = \{ ( \vv, \vh_{t}, \vz_{t}) \}$
   \STATE $\vg \leftarrow - \left( \nabla E ( \vx_0 ) + \sum_{t=1}^{\tau-1} \nabla E ( \vx_t ) - E ( \vy_{t-1} ) \right)$
   \STATE Run Algorithm \ref{alg:local_search}, and obtain $\tilde{\vv}_T, \tilde{\vh}_T^{(1)}, \tilde{\vh}_T^{(2)}$.
   \STATE Run a single Gibbs step from $\tilde{\vv}_T, \tilde{\vh}_T^{(1)}, \tilde{\vh}_T^{(2)}$, and obtain $\tilde{\vv}_{T+1}, \tilde{\vh}_{T+1}^{(1)}, \tilde{\vh}_{T+1}^{(2)}$.
   \STATE Set $\tilde{\vx}_0 = \tilde{\vy}_0 = ( \tilde{\vv}_{T+1}, \tilde{\vh}_{T+1}^{(1)}, \tilde{\vh}_{T+1}^{(2)} )$.
   \STATE Run Algorithm \ref{alg:maximal_coupling} , and obtain $\{ (\tilde{\vx}_{t}, \tilde{\vy}_{t-1}) \}, \ \tilde{\tau}$.
   \STATE $\vg \leftarrow \vg + \nabla E ( \tilde{\vx}_0 ) + \sum_{t=1}^{\tilde{\tau}-1} \nabla E ( \tilde{\vx}_t ) - E ( \tilde{\vy}_{t-1} )$
   \STATE $\vtheta \leftarrow \vtheta + \alpha \vg$
   \UNTIL{convergence}
   \STATE {\bfseries return} $\vtheta$
\end{algorithmic}
\end{algorithm}

\subsection{How to Apply to Non-binary Cases}
\label{subsec:binarization}
Because our UCD-LMI heavily relies on the local search algorithm, it is only applicable to the case where all the variable $\rvx$ is binary.
For example, when dealing with image data, it is common to use a DBM of a Gaussian-Bernoulli type~\citep{welling2004exponential,hinton2006reducing,liao2022gaussian}.
Because the visible units are defined as continuous variables in such cases, our method is difficult to be applied.
However, this problem can be avoided by binarizing data via preprocessing.
In the case of 8-bit RGB images, for instance, each pixel of an image takes an integer value from $0$ to $255$.
The integer value can be transformed into an $8$ dimensional binary vector by expressing it in the binary numerical system (e.g., $123 = 01111011_{(2)}$) without loss of information.
By using this binarization as preprocessing, we can directly apply our method to real-world data like natural images.

\section{Other Practical Techniques}
\label{sec:practical_tech}
In this section, we provide some additional techniques to successfully train DBMs using our UCD-LMI algorithm.
Basically, these techniques are designed to reduce the variance of the gradient estimates by our UCD-LMI, and accelerate its training.

\subsection{Simplified Centering Trick}
\label{subsec:centering}
In the original model definition, each of the visible and hidden units is assumed to take a value of $0$ or $1$.
In this case, a gradient of the energy function in terms of each element of the weights is the product of connected unit values as follows:
\begin{align}
    \nabla_{\emW_{i,j}^{(1)}} E (\vv, \vh^{(1)}, \vh^{(2)} ; \vtheta ) &= - \evv_{i} \evh^{(1)}_{j}, \label{eq:energy_gradient_1}\\
    \nabla_{\emW_{j,k}^{(2)}} E (\vv, \vh^{(1)}, \vh^{(2)} ; \vtheta ) &= - \evh^{(1)}_{j} \evh^{(2)}_{k}. \label{eq:energy_gradient_2}
\end{align}
This means that the gradient for $\emW_{i,j}^{(1)}$ is always $0$ except when both $\evv_i$ and $\evh^{(1)}_{j}$ are $1$, and the same applies to $\emW_{j,k}^{(2)}$.
For this reason, under the original model definition, the gradient tends to be sparse, leading to poor convergence.
The centering trick~\cite{montavon2012deep} is useful to alleviate the problem by normalizing the unit values using moving statistics.
However, using moving statistics introduces non-stationarity into the optimization, limiting the advantage of using unbiased MCMC.

To address this issue, we use a simpler strategy, in which each unit value of $\{ 0, 1 \}$ is cast into $\{ -1, 1 \}$ by mapping $0$ into $-1$ via preprocessing.
In this strategy, the gradients in Eqs. (\ref{eq:energy_gradient_1}) and (\ref{eq:energy_gradient_2}) always take nonzero values, alleviating the sparsity of the original definition without using moving statistics.
Note that when we adopt this definition, we need to modify the indicator function $\vone_{a \geq 0}$ in Algorithm \ref{alg:local_search} into a sign function, i.e., $\mathrm{sgn} \left( a \right) \coloneqq 2 \cdot \vone_{a \geq 0} - 1$.

\subsection{Variance Reduction by Marginalization}
\label{subsec:marginal}
Thanks to the bipartite structure of a DBM, marginal distributions over the even layers and the odd layers are both tractable as follows:
\begin{gather}
    P \left( \rvv, \rvh^{(2)} \right) \propto \exp \left(  - E_{\mathrm{even}} ( \rvv, \rvh^{(2)} ) \right), \\
    P \left( \rvh^{(1)} \right) \propto \exp \left(  - E_{\mathrm{odd}} ( \rvh^{(1)} ) \right), \\
    \begin{aligned}
        &E_{\mathrm{even}} ( \rvv, \rvh^{(2)} ) \\
        &= - \sum_j \log \cosh (\rvv^\top \mW_{:,j}^{(1)} + \mW_{j,:}^{(2)} \rvh^{(2)}),
    \end{aligned} \label{eq:marginal_energy_even}\\
    \begin{aligned}
        E_{\mathrm{odd}} ( \rvh^{(1)} ) = 
        &- \sum_i \log \cosh (\mW_{i,:}^{(1)} \rvh^{(1)}) \\
        &- \sum_k \log \cosh ({\rvh^{(1)}}^\top \mW_{k,:}^{(2)}),
    \end{aligned} \label{eq:marginal_energy_odd}
\end{gather}
where $\cosh$ is a hyperbolic cosine function, i.e., $\cosh (a) = (\exp(a) + \exp(-a)) / 2$.
Note that we here assume that each unit takes a value of $1$ or $-1$ as described in Section \ref{subsec:centering}.
Similarly, the marginal posterior distributions over the even and odd layers given the visible layer are also derived as follows:
\begin{gather}
    P \left( \rvh^{(2)} \mid \vv \right) \propto \exp \left(  - E_{\mathrm{even}} ( \rvv, \rvh^{(2)} ) \right), \\
    P \left( \rvh^{(1)} \mid \vv \right) \propto \exp \left(  - \tilde{E}_{\mathrm{odd}} ( \vv, \rvh^{(1)} ) \right), \\
    \begin{aligned}
        \tilde{E}_{\mathrm{odd}} ( \vv, \rvh^{(1)} ) = 
        &- \vv^\top \mW^{(1)} \rvh^{(1)} \\
        &- \sum_k \log \cosh ({\rvh^{(1)}}^\top \mW_{k,:}^{(2)}),
    \end{aligned}
\end{gather}
Based on these facts, Eq. (\ref{eq:gradient}) can be rewritten as follows:
\begin{align}
    &\nabla_\vtheta \log p ( \vv ; \vtheta ) \label{eq:marginalized_gradient}\\
    &\begin{aligned}
    = &- \frac{1}{2} \ \mathbb{E} \left[  \nabla_\vtheta ( E_{\mathrm{even}} (\vv, \rvh^{(2)} ; \vtheta ) +  \tilde{E}_{\mathrm{odd}} (\vv, \rvh^{(1)} ; \vtheta ) ) \right] \\
    &+ \frac{1}{2} \ \mathbb{E} \left[ \nabla_\vtheta ( E_{\mathrm{even}} (\tilde{\rvv}, \tilde{\rvh}^{(2)} ; \vtheta ) +  E_{\mathrm{odd}} ( \tilde{\rvh}^{(1)} ; \vtheta ) ) \right],
    \end{aligned} \nonumber
\end{align}
Using this expression for constructing an unbiased estimator with MH-based coupling as described before, we can even reduce its variance, keeping the estimator unbiased.

\subsection{Weight Initialization with Orthogonal Matrix}
\label{subsec:init}
Aside from the learning algorithm, initialization of the weight matrices is also a key to the successful training of a DBM.
A common way of initialization is to use random values chosen from a zero-mean Gaussian with a small variance (e.g., $0.01^2$)~\citep{hinton2012practical}.
However, It is more natural to change the scale of weight values depending on the dimensionality of the visible and hidden units.
In the context of initialization for FNNs, adaptively scaling methods proposed by \citet{glorot2010understanding} and \citet{he2015delving} are widely used, but these are not designed for DBMs.

As an alternative, we propose to initialize the weights of a DBM with random (semi-)orthogonal matrices.
Since the orthogonal matrix has a property that its inverse is equal to its transpose, i.e., $\mW^\top \mW = \mW \mW^\top = \mI$, 
it is a natural choice for a model with a bidirectional nature like a DBM.
Orthogonal initialization is often used for FNNs~\citep{saxe2013exact} also, and a way to create uniform random samples of orthogonal matrices is known~\citep{mezzadri2006generate}; hence we directly apply it to the case of DBMs.
How to initialize the bias parameters is provided in Appendix \ref{sec:bias_init}.

\subsection{Sampling in Test Time}
\label{subsec:sampling}
When we perform sampling from a trained DBM, a common way is to run Gibbs sampling starting from random noise for finite steps and treat the final state of the visible unit as a sample from the model.
However, we empirically find that it is difficult to obtain good samples in that way because Gibbs sampling is typically suffered from the curse of dimensionality.
Instead, as in the training phase, we first use the local search algorithm to find a local minimum of the energy function, and then perform the MH algorithm for a few steps to obtain the final state as a sample.
In practice, we can skip running the MH algorithm and just use the local minimum as a sample, because almost all proposals in the MH algorithm are rejected, and the sample hardly moves from the local minimum.


\section{Experiment}
In the experiments, we first verify that our MH-based coupling algorithm in the UCD-LMI is efficient compared to existing Gibbs-based coupling in the vanilla UCD using toy examples.
Subsequently, we validate the generative performance of DBMs trained with our method using MNIST and CIFAR-10 datasets, and compare it with existing training methods of DBMs and other deep generative models.


\begin{figure}
    \centering
    \begin{minipage}{0.49\hsize}
        \centering
        \includegraphics[width=\hsize]{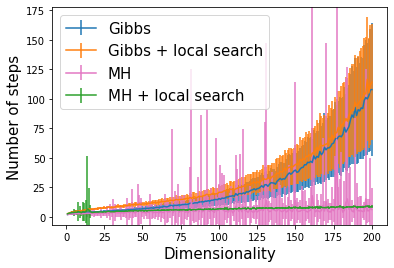}
    \end{minipage}
    \begin{minipage}{0.49\hsize}
        \centering
        \includegraphics[width=\hsize]{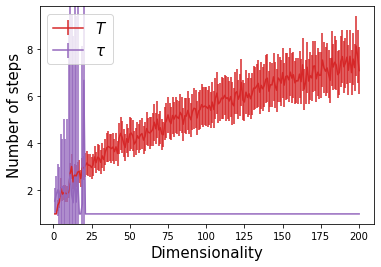}
    \end{minipage} \\
    \begin{minipage}{0.49\hsize}
        \centering
        (a) Coupling comparison
    \end{minipage}
    \begin{minipage}{0.49\hsize}
        \centering
        (b) Respective \# of steps
    \end{minipage}
    \caption{Efficiency comparison of coupling methods for unbiased gradient estimation. (a) We compare the total steps of local search and coupling (i.e., $\tau + T$ in Algorithms \ref{alg:maximal_coupling} and \ref{alg:local_search}) between Gibbs-based coupling and our MH-based coupling when changing the dimensionality from $1$ to $200$. (b) We also provide the respective number of steps of $T$ and $\tau$ for our MH-based coupling.}
    \label{fig:coupling_comparison}
\end{figure}

\subsection{Toy Examples}
\label{subsec:toy}
In this experiment, we prepare randomly initialized RBMs, i.e., single-layer DBMs, and run some coupling algorithms to estimate the gradient of its partition function, i.e., the second term of Eq. (\ref{eq:gradient}), to compare their efficiency under various conditions.
We change the dimensionality of the visible units and the hidden units in the range from $1$ to $200$.
The weights of RBMs are initialized with random orthogonal matrices.

We compare our MH-based method with Gibbs-based maximal coupling proposed by \citet{Qiu2020Unbiased}.
When performing Gibbs-based coupling, we simply initialize the state with uniform noise.
To investigate the effect of the initialization by local search, we also try the cases where Gibbs-based coupling is initialized by local search and MH-based coupling is initialized by uniform noise.
For a fair comparison, we compare the total time steps it takes to obtain the estimator, including the steps of local search.

The results are shown in Figure \ref{fig:coupling_comparison}.
We observe that the coupling time of the Gibbs-based method grows exponentially with dimensionality.
In the training of DBMs for real-world data (e.g., images), we need to handle much more high-dimensional data; hence these results show that the Gibbs-based method is no longer practical for such cases.
Unfortunately, initializing Gibbs-based coupling with a local mode does not help to reduce the coupling time.

On the other hand, it can be seen that our MH-based method works efficiently even for high-dimensional cases.
When the state is initialized with uniform noise, the variance of $\tau + T$ is very high; hence the local mode initialization (LMI) contributes to variance reduction.
We also observe that the coupling time $\tau$ is always $1$ for MH-based coupling with the LMI when $d \geq 25$.
Although the number of local search steps also increases with dimensionality, the increase is gradual compared to the Gibbs-based coupling, and the variance is also much smaller in high dimensions.
These results demonstrate the efficiency of our coupling method to be practically used for the gradient estimator in the training of DBMs for high-dimensional data (e.g., natural images).

\begin{table}[t]
    \centering
    \caption{Quantitative results of the image generation for MNIST and CIFAR-10. We report the Fréchet Inception Distance (FID) score. Lower is better.}
    \begin{tabular}{lc}
        \toprule
        Model                                           &FID \\ \midrule
        {\bf MNIST} & \\ \midrule
        VAE                                             &$16.14$ \\
        2sVAE~\citep{dai2019diagnosing}                 &$12.60$ \\
        PixelCNN++~\citep{salimans2017pixelcnn++}       &$11.38$ \\
        WGAN~\citep{arjovsky2017wasserstein}            &$10.28$ \\
        NVAE~\citep{vahdat2020nvae}                     &$7.93$ \\
        GB-RBM~\citep{liao2022gaussian} (Gibbs)         &$47.53$ \\
        GB-RBM~\citep{liao2022gaussian} (Gibbs-Langevin)&$17.49$ \\
        Centered DBM~\citep{montavon2012deep}           &$89.10$ \\
        {\bf Unbiased DBM (ours)}                       &$10.33$ \\ \midrule
        {\bf CIFAR-10} & \\ \midrule
        PixelCNN~\citep{van2016pixel}                   &$65.93$ \\
        DCGAN~\citep{radford2015unsupervised}           &$37.11$ \\
        WGAN-GP~\citep{gulrajani2017improved}           &$36.40$ \\
        {\bf Unbiased DBM (ours)}                       &$36.58$ \\
        \bottomrule
    \end{tabular}
    \label{tab:fid}
\end{table}

\begin{table*}[t]
    \centering
    \caption{Ablation study using the MNIST dataset.}
    \begin{tabular}{ccccc}
        \toprule
        \multicolumn{4}{c}{Configuration}                            & \\ \cmidrule{1-4}
        Unbiased gradient    &Simplified centering & Marginalization & Orthogonal initialization &FID \\ \midrule
        \checkmark           &\checkmark &\checkmark       &\checkmark        & $10.33$\\ 
                             &\checkmark &\checkmark       &\checkmark        & $112.1$\\ 
        \checkmark           &           &\checkmark       &\checkmark        & $30.27$\\ 
        \checkmark           &\checkmark &                 &\checkmark        & $13.70$\\ 
        \checkmark           &\checkmark &\checkmark       &                  & $18.20$\\ 
        \bottomrule
    \end{tabular}
    \label{tab:ablation}
\end{table*}
\subsection{Image Generation}
\label{subsec:image_gen}
We train DBMs for image datasets including MNIST~\citep{deng2012mnist}, Fashin-MNIST~\citep{xiao2017fashion} and CIFAR-10.
We do not use greedy layer-wise pretraining, which we find that is not necessary for our UCD-LMI algorithm.
The implementation details are provided in Appendix \ref{sec:impl}.
We compare their performance with other existing deep generative models (DGMs), including variational autoencoders (VAEs), generative adversarial networks (GANs), and autoregressive models.
For MNIST, We also compare our method with the centered DBM~\citep{montavon2012deep} as another way to train a DBM without pretraining.
Unfortunately, the original UCD algorithm does not work for this setting because its coupling time is extremely long; hence we omit it from the comparison.
On the other hand, we observe that the coupling time of our UCD-LMI falls within the range from 5 to 30 for all datasets, showing that our algorithm can withstand practical use.

\begin{figure}[tb]
    \centering
    MNIST \\
    \begin{minipage}{0.49\linewidth}
        \centering
        \includegraphics[width=\hsize]{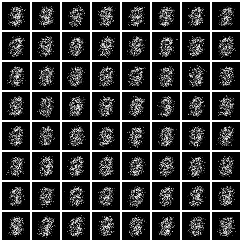}
    \end{minipage}
    \begin{minipage}{0.49\linewidth}
        \centering
        \includegraphics[width=\hsize]{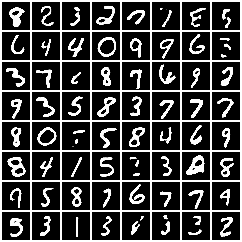}
    \end{minipage} \\
    \begin{minipage}{\linewidth}
        \centering
    \end{minipage}
    \begin{minipage}{\linewidth}
        \centering
        Fashion MNIST
    \end{minipage} \\
    \begin{minipage}{0.49\linewidth}
        \centering
        \includegraphics[width=\hsize]{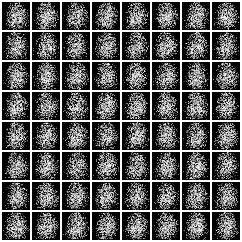}
    \end{minipage}
    \begin{minipage}{0.49\linewidth}
        \centering
        \includegraphics[width=\hsize]{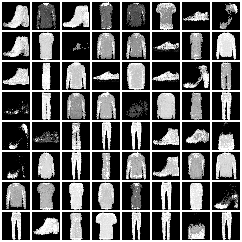}
    \end{minipage}
    \begin{minipage}{0.49\linewidth}
        \centering
        (a) Baseline DBM (PCD)
    \end{minipage}
    \begin{minipage}{0.49\linewidth}
        \centering
        (b) Unbiased DBM (ours)
    \end{minipage}
    \caption{Generated samples of MNIST and Fashion-MNIST by a DBM trained with (a) a baseline method using PCD and (b) our UCD-LMI algorithm.}
    \label{fig:mnist_sample}
\end{figure}


\begin{figure}[tb]
    \centering
    \begin{minipage}{0.49\hsize}
        \centering
        Ground truth \\
        \includegraphics[width=\hsize]{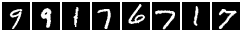} \\
        Masked images \\
        \includegraphics[width=\hsize]{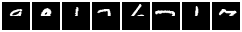} \\
        Completion results \\
        \includegraphics[width=\hsize]{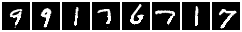} \\
    \end{minipage}
    \begin{minipage}{0.49\hsize}
        \centering
        Ground truth \\
        \includegraphics[width=\hsize]{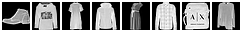} \\
        Masked images \\
        \includegraphics[width=\hsize]{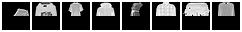} \\
        Completion results \\
        \includegraphics[width=\hsize]{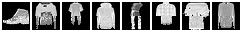} \\
    \end{minipage}
    \caption{Completion}
    \label{fig:completion}
\end{figure}


The quantitative result using the Fréchet Inception Distance (FID) score for MNIST and CIFAR-10 is summarized in Table \ref{tab:fid}.
Our method is labeled as "Unbiased DBM" in the table.
It can be seen that our unbiased DBM achieves FID scores comparable to other DGMs, such as WGAN and WGAN-GP.
For MNIST, we also provide the ablation study in Table \ref{tab:ablation} to demonstrate the effectiveness of each component of our method, including the unbiased gradient estimation, the simplified centering trick, the variance reduction by marginalization, and the orthogonal initialization.
When we train DBMs without the unbiased gradient estimator, the generative performance drastically drops, demonstrating that it is crucial for proper training of DBMs without greedy layer-wise pretraining.
It can be observed that other techniques are also effective, while the improvement in FID scores is not so great compared to the importance of the unbiased gradient.

Figure \ref{fig:mnist_sample} shows a comparison of generated samples of MNIST and Fashion-MNIST by trained DBMs using the vanilla PCD and our UCD-LMI without greedy pretraining.
While the DBM trained with the vanilla PCD completely fails to learn data structure and generates meaningless samples, our UCD-LMI successfully generates clean samples.

One of the advantages of DBMs over other DGMs is that DBMs can also be used for missing value imputation of incomplete data.
Figure \ref{fig:completion} shows examples of completed images of MNIST and Fashion-MNIST by a trained DBM with our UCD-LMI.
We create incomplete images by masking the lower half pixels of the original images.
When performing the completion, we run the local search algorithm to find a low-energy state given the incomplete data, and fill in the missing values using the obtained low-energy state.
It can be seen that the completed images are close to but not exactly the same as the ground truth because of the uncertainty.


\section{Conclusion}
\label{sec:discussion}
In this paper, we revisit the difficulty in training deep Boltzmann machines (DBMs), and propose a practical algorithm to obtain unbiased gradient estimates using the coupling technique based on the Metropolis–Hastings and the local search (Section \ref{sec:method}). 
We also provide some practical techniques to successfully train a DBM (Section \ref{sec:practical_tech}). 

In the empirical study, we first demonstrate that our MH-based coupling is much more efficient than the existing Gibbs-based coupling (Section \ref{subsec:toy}). 
We also perform experiments of image generation, and show that DBMs trained with our method achieves comparable performance to other DGMs in terms of the FID score (Section \ref{subsec:image_gen}).

One of the remaining challenges of the DBM is that it is still difficult to be applied for extremely high-dimensional inputs like high-resolution images, because the weights are fully connected between layers, resulting in the growth of the number of parameters.
By applying our UCD-LMI algorithm to locally-connected DBMs (e.g., convolutional Boltzmann machines~\citep{lee2009convolutional}), the problem could be alleviated, which is an important direction of future work.

\section*{Acknowledgements}
This work was supported by JSPS KAKENHI Grant Number JP21J22342 and the Mohammed bin Salman Center for Future Science and Technology for Saudi-Japan Vision 2030 at The University of Tokyo (MbSC2030). 
Computational resources of AI Bridging Cloud Infrastructure (ABCI) provided by National Institute of Advanced Industrial Science and Technology (AIST) were used for the experiments.

\bibliography{example_paper}
\bibliographystyle{icml2023}

\newpage
\appendix
\onecolumn

\begin{algorithm}[tb]
   \caption{Maximal Coupling of Metropolis-Hastings Algorithm with Uniform Proposal}
   \label{alg:maximal_coupling}
\begin{algorithmic}
   \STATE Draw $\vx^\prime \sim \mathrm{Unf} ( \vx^\prime )$.
   \STATE $A_x \leftarrow \min \left( 1, \exp ( E ( \vx_0 ; \vtheta ) - E ( \vx^\prime ; \vtheta ) ) \right)$
   \STATE Draw $u$ uniformly from $[0, 1)$.
   \STATE {\bfseries if} $u < A_x$ {\bfseries then} $\vx_1 \leftarrow \vx^\prime$ {\bfseries else} $\vx_1 \leftarrow \vx_0$
   \STATE $t \leftarrow 1$
   \REPEAT
   \STATE Draw $\vx^\prime \sim \mathrm{Unf} ( \vx^\prime )$.
   \STATE $A_x \leftarrow \min \left( 1, \exp ( E ( \vx_{t} ; \vtheta ) - E ( \vx^\prime ; \vtheta ) ) \right)$
   \STATE $A_y \leftarrow \min \left( 1, \exp ( E ( \vy_{t-1} ; \vtheta ) - E ( \vx^\prime ; \vtheta ) ) \right)$
   \STATE Draw $u$ uniformly from $[0, 1)$.
   \STATE {\bfseries if} $u < A_x$ {\bfseries then} $\vx_{t+1} \leftarrow \vx^\prime$ {\bfseries else} $\vx_{t+1} \leftarrow \vx_{t}$
   \STATE {\bfseries if} $u < A_y$ {\bfseries then} $\vy_{t} \leftarrow \vx^\prime$ {\bfseries else} $\vy_t \leftarrow \vy_{t-1}$
   \STATE $t \leftarrow t + 1$
   \UNTIL{$\vx_t = \vy_{t-1}$}
   \STATE $\tau \leftarrow t$
   \STATE {\bfseries return} $\{ (\vx_{t}, \vy_{t}) \}$, $\tau$
\end{algorithmic}
\end{algorithm}

\begin{algorithm}[tb]
   \caption{Maximal Coupling of Metropolis–Hastings Algorithm for Posterior $P (\vh \mid \vv )$}
   \label{alg:maximal_coupling_for_posterior}
\begin{algorithmic}
   \STATE Draw $\vx^\prime \sim \mathrm{Unf} ( \vh^\prime )$.
   \STATE $A_h \leftarrow \min \left( 1, \exp ( E ( \vv, \vh_0 ; \vtheta ) - E ( \vv, \vh^\prime ; \vtheta ) ) \right)$
   \STATE Draw $u$ uniformly from $[0, 1)$.
   \STATE {\bfseries if} $u < A_h$ {\bfseries then} $\vh_1 \leftarrow \vh^\prime$ {\bfseries else} $\vh_1 \leftarrow \vh_0$
   \STATE $t \leftarrow 1$
   \REPEAT
   \STATE Draw $\vh^\prime \sim \mathrm{Unf} ( \vh^\prime )$.
   \STATE $A_h \leftarrow \min \left( 1, \exp ( E ( \vv, \vh_{t} ; \vtheta ) - E ( \vv, \vh^\prime ; \vtheta ) ) \right)$
   \STATE $A_z \leftarrow \min \left( 1, \exp ( E ( \vz_{t-1} ; \vtheta ) - E ( \vh^\prime ; \vtheta ) ) \right)$
   \STATE Draw $u$ uniformly from $[0, 1)$.
   \STATE {\bfseries if} $u < A_h$ {\bfseries then} $\vh_{t+1} \leftarrow \vh^\prime$ {\bfseries else} $\vh_{t+1} \leftarrow \vh_{t}$
   \STATE {\bfseries if} $u < A_z$ {\bfseries then} $\vz_{t} \leftarrow \vh^\prime$ {\bfseries else} $\vz_t \leftarrow \vz_{t-1}$
   \STATE $t \leftarrow t + 1$
   \UNTIL{$\vh_t = \vz_{t-1}$}
   \STATE $\tau \leftarrow t$
   \STATE {\bfseries return} $\{ (\vh_{t}, \vz_{t}) \}$, $\tau$
\end{algorithmic}
\end{algorithm}

\begin{algorithm}[tb]
   \caption{Local Search Algorithm for Posterior $P \left( \vh \mid \vv \right)$}
   \label{alg:local_search_for_posterior}
\begin{algorithmic}
   \STATE Draw an initial state $\left( \vh_0^{(1)}, \vh_0^{(2)} \right)$ uniformly.
   \STATE Draw $u$ uniformly from $[0, 1)$. 
   \STATE $t \leftarrow 0$
   \REPEAT
   \IF{$u < 0.5$}
   \STATE $\vh^{(2)}_{t+1} = \vone_{{\vh_t^{(1)}}^\top \mW^{(2)} \geq 0}$
   \STATE $\vh^{(1)}_{t+1} = \vone_{\vv^\top \mW^{(1)} + \mW^{(2)} \vh_{t+1}^{(2)} \geq 0}$
   \ELSE
   \STATE $\vh_{t+1}^{(1)} = \vone_{\vv^\top \mW^{(1)} + \mW^{(2)} \vh_t^{(2)} \geq 0}$
   \STATE $\vh_{t+1}^{(2)} \leftarrow \vone_{{\vh_{t+1}^{(1)}}^\top \mW^{(2)} \geq 0}$
   \ENDIF
   \STATE $t \leftarrow t + 1$
   \UNTIL{$\left( \vh_t^{(1)}, \vh_t^{(2)} \right) = \left( \vh_{t-1}^{(1)}, \vh_{t-1}^{(2)} \right)$}
   \STATE $T \leftarrow t$
   \STATE {\bfseries return} $\left( \vh_T^{(1)}, \vh_T^{(2)} \right)$, $T$
\end{algorithmic}
\end{algorithm}

\section{Unbiased Estimator for the First Term of Eq. (\ref{eq:gradient})}
\label{sec:unbiased_first_term}
Similarly to Eq. (\ref{eq:unbiased_energy_grad}), the unbiased estimator for the first term of Eq. (\ref{eq:gradient}) can be constructed as follows:
\begin{align}
    &\mathbb{E} \left[ \nabla E ( \vv, \rvh ; \vtheta ) \right]  \label{eq:unbiased_positive_energy_grad}\\
    &= \mathbb{E} \left[ \nabla E ( \vv, \rvh_0 ; \vtheta ) + \sum_{t=1}^{\tau-1} \nabla E ( \vv, \rvh_t ; \vtheta ) - \nabla E ( \vv, \rvz_{t-1} ; \vtheta ) \right], \nonumber
\end{align}
, where $\rvh = ( \rvh^{(1)} , \rvh^{(2} )$, and $\{ (\rvh_t, \rvz_t) \}$ is a coupling of Markov chains that satisfies (1) $\mathbb{E} \left[ \nabla E \left( \vv, \rvh_t; \vtheta \right) \right] \to \mathbb{E} \left[ \nabla E \left( \vv, \rvh; \vtheta \right) \right]$, (2) $\rvh_t$ and $\rvz_t$ have the same marginal distributions for all $t \geq 0$, and (3) $\rvh_t = \rvz_{t-1}$ after some random time $\tau$.
A maximal coupling based on Metroplis-Hastings algorithm for this estimator is provided in Algorithm \ref{alg:maximal_coupling_for_posterior}.
To initialize this coupling with a state around a local mode of the posterior $P ( \rvh \mid \vv )$, we use the local search as described in Algorithm \ref{alg:local_search_for_posterior}.

\section{Initialization of Bias Parameters}
\label{sec:bias_init}
We empirically observe that the initialization of bias parameters is also important especially when we use the simplified centering trick in Section \ref{subsec:centering}.
A common way to initialize the bias parameters is simply setting $0$.
However, if we use the centering trick and set the bias zero, the energy function will be symmetric for the unit flip, i.e. $E ( \vv, \vh ; \vtheta ) = E ( -\vv, -\vh ; \vtheta )$.
To avoid this, we initialize the bias parameters by sampling from a logistic distribution $\mathrm{Logistic} (0, 0.5)$, which we find works well in practice.

\section{Implementation Details}
\label{sec:impl}
In the experiment of image generation, we set the number of hidden units per layer equal to the visible units for MNIST and Fashion-MNIST, which means it is set to $6,272 \ (= 1 \times 28^2 \times 8)$.
For CIFAR-10, we set it to $10,000$.
We use the SGD as an optimizer and set the learning rate to $1 \times 10^{-2}$.
We run $100,000$ gradient steps of training for both datasets.
The implementation is available at \url{https://github.com/iShohei220/unbiased_dbm}.

\nocite{langley00}


\end{document}